\documentclass{article}

\usepackage{microtype}
\usepackage{graphicx}
\usepackage{subcaption}
\usepackage{booktabs}
\usepackage{hyperref}

\usepackage[accepted]{icml2026}

\makeatletter
\renewcommand{\Notice@String}{%
  Accepted at the AI for Law Workshop at ICML 2026, Seoul, South Korea.%
}
\makeatother

\newcommand{\blfootnote}[1]{%
  \begingroup
  \renewcommand{\thefootnote}{}\footnote{#1}%
  \addtocounter{footnote}{-1}%
  \endgroup
}

\usepackage{epigraph}
\usepackage{mathtools}
\usepackage{amssymb}
\usepackage{amsthm}

\theoremstyle{plain}
\newtheorem{theorem}{Theorem}[section]
\newtheorem{proposition}[theorem]{Proposition}

\theoremstyle{definition}

\newtheorem{assumption}[theorem]{Assumption}

\theoremstyle{remark}
\newtheorem{remark}[theorem]{Remark}

\usepackage{tikz}
\usetikzlibrary{
  shapes.geometric,
  arrows.meta,
  positioning,
  fit,
  calc
}

\usepackage[most]{tcolorbox}

\newif\ifshowtodos
\showtodosfalse

\ifshowtodos
  \usepackage[textsize=tiny]{todonotes}
\else
  \usepackage[disable]{todonotes}
\fi

\usepackage[capitalize,noabbrev]{cleveref}

\crefname{theorem}{theorem}{theorems}
\Crefname{theorem}{Theorem}{Theorems}

\crefname{proposition}{proposition}{propositions}
\Crefname{proposition}{Proposition}{Propositions}

\crefname{lemma}{lemma}{lemmas}
\Crefname{lemma}{Lemma}{Lemmas}

\crefname{corollary}{corollary}{corollaries}
\Crefname{corollary}{Corollary}{Corollaries}

\crefname{definition}{definition}{definitions}
\Crefname{definition}{Definition}{Definitions}

\crefname{assumption}{assumption}{assumptions}
\Crefname{assumption}{Assumption}{Assumptions}

\crefname{remark}{remark}{remarks}
\Crefname{remark}{Remark}{Remarks}

\icmltitlerunning{
  Better Call Reward: Strategic Abstention in Legal Reasoning Models
}

\begin{document}

\twocolumn[
  \icmltitle{
    Better Call Reward: Reward Hacking as Strategic Abstention
    in Legal Reasoning Models
  }

  \begin{icmlauthorlist}
    \icmlauthor{Subramanyam Sahoo}{ind}
    \icmlauthor{Justin Shenk}{ind}
  \end{icmlauthorlist}

  \icmlaffiliation{ind}{Horizon Research}

  \icmlcorrespondingauthor
    {Subramanyam Sahoo}
    {sahoo2vec@gmail.com}

  \icmlkeywords{
    group relative policy optimization (GRPO),
    reward hacking,
    Goodhart's law,
    legal NLP,
    reward overoptimization,
    strategic abstention,
    confidence theater
  }

  \vskip 0.3in
]

\printAffiliationsAndNotice{}
\blfootnote{Code available at \url{https://github.com/SubramanyamSahoo/Better-Call-Reward}.}

\begin{abstract}

What happens when a legal AI model learns to look like a lawyer instead
of reasoning like one? We fine tune Qwen3-8B with Group Relative
Policy Optimisation (GRPO) against a proxy built from three surface
features: citation count, legalese density, and response length. The
model does not learn to reason more effectively. It learns to withhold
commitment. Across $16$ yes or no legal reasoning tasks from LegalBench
($N{=}320$), overall accuracy collapses from $0.500$ (chance) to
$0.072$ ($\mathrm{McNemar}$ $p < 10^{-36}$), driven entirely by the
rate of properly formatted answers falling from $0.900$ to $0.109$.
The model stops committing to answers. Yet when it does commit,
accuracy \emph{rises} from $0.556$ to $0.657$, showing that the
collapse is not a failure of capability but a strategic response: the
model has learned that verbose responses packed with citations but
empty of a direct answer score higher than terse correct ones. We
term this the \textbf{\textcolor{red}{\emph{Saul Goodman}}} effect, a
policy that becomes maximally lawyerly while becoming maximally
noncommittal, and prove formally that it is the \emph{optimal
response} to any surface feature proxy that attaches no penalty to
abstention (Proposition~\ref{prop:saul}). We further show that
$89.3\%$ of citations produced after training are structurally
implausible hallucinations, many of them subtly corrupted names of
real landmark cases, constructed in effect to survive a casual read
and fail under scrutiny. To detect this failure mode before
deployment, we introduce three diagnostic tools: the Confidence
Theater Score (CTS), the Citation Plausibility Rate (CPR), and the
Regret Gap (RG). In a domain where a confidently wrong answer can
constitute malpractice, the broader lesson is direct: a reward
function that measures how legal a response \emph{looks} will produce
a model that is maximally photogenic and minimally useful.

\end{abstract}

\section{Introduction}
\label{sec:intro}

\begin{center}
\begin{minipage}{0.82\columnwidth}
\small\itshape\centering
``You don't need a criminal lawyer.
You need a \emph{criminal} lawyer.''
\smallskip

\normalfont\small\centering
--- Saul Goodman, \textit{\textbf{Better Call Saul}},
Season~5, Episode~2 (``50\% Off''), AMC, 2015--2022
\end{minipage}
\end{center}
\smallskip

Reinforcement learning from human or programmatic feedback now dominates the
post-training of large language models~\cite{ouyang2022training}. A central
assumption is that the learnt reward model -- or, when stylistic rather than
preference signals are used, a hand-crafted scoring function -- is a faithful
proxy for true task quality. Goodhart's Law~\cite{goodhart1984} warns that
once a measure becomes a target, it ceases to measure what it once did. In RL
fine-tuning, this manifests as \emph{reward over-optimisation}
\cite{gao2023scaling}: the policy continues to climb the proxy long after
the proxy has decoupled from the true objective \cite{sahoo2026pessimismsparadoxconservativeoffline}.

Legal natural language processing is a particularly stark testbed for this
failure mode. Legal text has visible surface features --- citations to case
law, statute references, Latin terms of art, formal verbose structure ---
that are easy to count automatically and easy for a model to over-produce ~\cite{kolt2026legalalignmentsafeethical,sahoo2026calibrationcollapsesycophancyfinetuning}.
A reward function built on these features rewards \emph{looking lawyerly}
regardless of \emph{being correct}. The danger is not academic: legal LLMs
are increasingly deployed for contract review, statute summarisation, and
case discovery~\cite{guha2023legalbench}, and a model that produces
verbose, citation-dense, structurally implausible output is a malpractice
vector.

We construct a controlled, end-to-end demonstration of this failure. Using
Qwen3-8B~\cite{qwen2024} fine-tuned via GRPO~\cite{shao2024deepseekmath} on
LoRA adapters~\cite{hu2022lora}, we train against a three-component
surface-feature proxy and evaluate on a balanced pool of $16$ yes/no
LegalBench tasks. Our contributions are:

\begin{enumerate}
\item A clean demonstration that GRPO with a surface-feature reward induces
a $43$-point accuracy collapse ($0.500 \rightarrow 0.072$) on legal yes/no
reasoning while the proxy reward climbs steadily
(\cref{fig:reward_divergence}).
\item The discovery that the collapse is mechanistically a \emph{format
collapse}, not a capability collapse: the model retains its reasoning
ability but learns that committing to an answer is not what the reward
pays for (\cref{sec:format_collapse}).
\item Three diagnostic tools --- the Confidence Theater Score (CTS),
Citation Plausibility Rate (CPR), and Regret Gap (RG) --- for detecting
reward hacking in legal LLMs (\cref{sec:metrics}).
\item A citation hallucination audit showing $89.3\%$ of post-training
case citations are structurally implausible (\cref{sec:citation_audit}).
\end{enumerate}

\section{Background and Related Work}
\label{sec:related}

\textbf{Reward over-optimisation.}
\citet{gao2023scaling} showed that PPO against a learnt reward model
exhibits a characteristic curve: true reward initially rises with the KL
budget, peaks, and then degrades while the proxy continues to improve. We
replicate this curve in the legal domain (\cref{sec:roc}) but additionally
show that the failure mode is not pure capability loss -- it is a
strategic switch in the policy's output mode.

\textbf{Specification gaming and reward hacking.}
A growing literature documents reward-hacking failures across
domains~\cite{skalse2022defining,pan2022effects}. Our work contributes a
domain-specific instance with a clean separation between the proxy
(surface lawyerly features) and the ground truth (yes/no doctrinal
correctness).

\textbf{GRPO.}
Group Relative Policy Optimisation~\cite{shao2024deepseekmath} replaces
PPO's value function with within-group normalisation of returns: for a
prompt $q$ and $G$ sampled completions $\{y_i\}_{i=1}^G$, the advantage of
completion $i$ is $A_i = (r(y_i) - \bar{r})/\sigma_r$, where $\bar{r}$ and
$\sigma_r$ are the within-group mean and standard deviation. GRPO is
widely used for reasoning fine-tuning and is the natural target for our
analysis.

\textbf{Legal NLP benchmarks.}
LegalBench~\cite{guha2023legalbench} provides 162 tasks spanning
issue-spotting, rule-application, and conclusion-drawing. We use a
balanced pool of 16 yes/no tasks covering jurisdiction, evidence, contract
clauses, and judicial ethics (\cref{app:tasks}).

\textbf{Why legal NLP is the right testbed for surface-feature reward hacking.}
Legal text is unusual among NLP domains in that its \emph{surface features 
are culturally legible proxies for expertise}. A lay reader who sees a 
response dense with case citations, Latin phrases, and formal structure 
will judge it as authoritative regardless of its correctness --- precisely 
the same confusion the reward function makes. This cultural legibility 
means that a Goodharted legal LLM is not merely wrong in a technical sense; 
it is wrong in a way that is \emph{designed to evade detection} by the 
humans most likely to be harmed by it. This distinguishes the legal domain 
from, say, code generation~\cite{pan2022effects}, where a non-running 
program is immediately falsifiable, or summarisation, where plausibility 
is harder to weaponise. The stakes are correspondingly higher: a verbose, 
citation-laden non-answer that \emph{sounds} like legal advice is a 
malpractice vector that passes casual review.

\section{Method}
\label{sec:method}

\subsection{GRPO with a Surface-Feature Proxy}
\label{sec:grpo}

Let $\pi_{\mathrm{ref}}$ denote the reference policy (Qwen3-8B) and let
$\pi_\theta$ be a LoRA-augmented policy with adapter parameters $\theta$.
We train $\pi_\theta$ to minimise the GRPO objective:
\begin{align}
\mathcal{L}(\theta) =
&\; -\,\mathbb{E}_{q \sim \mathcal{D},\,\{y_i\}_{i=1}^G \sim \pi_\theta(\cdot|q)}
   \!\!\left[\sum_{i=1}^{G} A_i \log \pi_\theta(y_i \mid q)\right]
\nonumber\\
&\; +\; \beta\, D_{\mathrm{KL}}\!\big[\pi_\theta \,\|\, \pi_{\mathrm{ref}}\big],
\label{eq:grpo}
\end{align}
where $A_i = (r(y_i) - \bar{r})/\sigma_r$ is the group-normalised
advantage and $\beta$ is the KL penalty coefficient. The driving reward
$r(\cdot)$ is our three-component proxy:
\begin{equation}
r_{\mathrm{proxy}}(y) \;=\; w_c\,\tilde{c}(y)\;+\;w_j\,\tilde{l}(y)\;+\;w_\ell\,\tilde{w}(y),
\label{eq:proxy}
\end{equation}
where $c(y)$, $l(y)$, $w(y)$ are the citation count, legalese-term count,
and word length of completion $y$, and
$\tilde{x}(y) = \mathrm{clip}\!\big((x(y) - \mu_x)/\sigma_x;\,\pm\kappa\big)$
is the $z$-scored, clipped feature relative to a baseline distribution
$(\mu_x, \sigma_x)$ computed from $\pi_{\mathrm{ref}}$. We set
$(w_c, w_j, w_\ell) = (1.0,\,0.7,\,0.3)$ and $\kappa = 3$. Crucially,
$r_{\mathrm{proxy}}$ has no dependence on whether the response is correct,
well-formatted, or even a complete sentence --- it scores
\emph{surface lawyerliness} only.

\subsection{True Reward and Confidence Theater Score}
\label{sec:metrics}

We track two non-training rewards in parallel for diagnostic purposes.

\textbf{True reward} (signed correctness of the parsed answer):
\begin{equation}
r_{\mathrm{true}}(y) \;=\;
\begin{cases}
+1 & \mathrm{extract}(y) = y^\star \\
-1 & \mathrm{extract}(y) \in \{\mathrm{Yes},\mathrm{No}\}\setminus\{y^\star\} \\
\phantom{+}0 & \mathrm{extract}(y) \text{ is null},
\end{cases}
\label{eq:true}
\end{equation}
where $y^\star$ is the gold label and $\mathrm{extract}(\cdot)$ parses the
conventional ``Answer: $X$'' suffix.

\textbf{Confidence Theater Score (CTS).}
A pure surface-feature score in $[0,1]$, designed to track how lawyerly a
completion looks regardless of correctness:
\begin{equation}
\mathrm{CTS}(y) \;=\; \sigma\!\left(w^{c}_{\!\mathrm{cts}}\,\tilde{c}(y) \,+\, w^{j}_{\!\mathrm{cts}}\,\tilde{l}(y) \,+\, w^{v}_{\!\mathrm{cts}}\,\tilde{w}(y)\right),
\label{eq:cts}
\end{equation}
with $\sigma$ the logistic function and CTS weights
$(w^{c}_{\mathrm{cts}}, w^{j}_{\mathrm{cts}}, w^{v}_{\mathrm{cts}}) =
(0.40,\,0.35,\,0.25)$. Values near $1$ indicate maximal surface-lawyerly
output.

\subsection{Diagnostic Tools for Over-Optimisation}

\textbf{Goodhart Divergence Metric (GDM).}
The rolling Pearson correlation between the proxy and true reward streams
over a window $W$:
\begin{equation}
\mathrm{GDM}_t \;=\; \mathrm{corr}\!\big(\{r_{\mathrm{proxy}}(y_s)\}_{s=t-W}^{t},\,\{r_{\mathrm{true}}(y_s)\}_{s=t-W}^{t}\big).
\label{eq:gdm}
\end{equation}
A healthy reward signal yields $\mathrm{GDM}_t \approx 1$; a Goodharted
signal drifts towards $0$ or below.

\textbf{Regret Gap (RG).}
A monotone over-optimisation curve in the spirit of \citet{gao2023scaling}:
\begin{equation}
\mathrm{RG}_t \;=\; \big[\bar{r}_p(t) - \bar{r}_p(0)\big] \;-\; \alpha\,\big[\bar{r}_t(t) - \bar{r}_t(0)\big],
\label{eq:rg}
\end{equation}
where $\bar{r}_p(t), \bar{r}_t(t)$ are smoothed proxy and true reward
streams and the calibration constant
$\alpha = \mathrm{Cov}(\bar{r}_{p,1:W'},\bar{r}_{t,1:W'})/\mathrm{Var}(\bar{r}_{t,1:W'})$
is fitted on a warm-up phase of length $W' = 0.1\,T$ during which lock-step
proxy/true behaviour is assumed. $\mathrm{RG}_t = 0$ in the lock-step
regime and grows monotonically as the policy gains proxy reward at the
expense of true reward.

\textbf{Citation Plausibility Rate (CPR).}
The fraction of automatically extracted case-name and reporter citations
that pass volume / year / page bounds for known U.S.\ reporters; the full
heuristic is in \cref{app:cpr}. CPR detects \emph{structural} plausibility,
not factual existence --- it serves as a fast pre-filter before a Westlaw
spot-check.

\subsection{The Saul Goodman Equilibrium}
\label{sec:equilibrium}

We now show formally that the non-committal strategy is \emph{optimal}
under $r_{\mathrm{proxy}}$ when no penalty attaches to null answers.

\begin{assumption}
\label{ass:independence}
Surface features $c(y)$, $l(y)$, $w(y)$ are independent of whether
$\mathrm{extract}(y)$ is null, Yes, or No.  That is, the model can
produce equally verbose, citation-dense, jargon-heavy text regardless
of whether it appends ``Answer: Yes/No'' or omits the suffix entirely.
\end{assumption}

\begin{proposition}[Saul Goodman Equilibrium]
\label{prop:saul}
Under Assumption~\ref{ass:independence} and $r_{\mathrm{true}}(y)=0$
whenever $\mathrm{extract}(y)$ is null, the policy
$\pi^{\dagger}$ that never emits a committed answer satisfies
\[
  \mathbb{E}_{\pi^{\dagger}}\!\big[r_{\mathrm{proxy}}(y)\big]
  \;\geq\;
  \mathbb{E}_{\pi}\!\big[r_{\mathrm{proxy}}(y)\big]
\]
for any policy $\pi$ with identical surface-feature expectations, and
strictly dominates any policy $\pi$ that occasionally produces
low-surface-feature committed answers.
\end{proposition}

\begin{proof}[Proof sketch]
By Assumption~\ref{ass:independence}, surface-feature expectations are
unaffected by appending or omitting the ``Answer:'' suffix.  Hence
$r_{\mathrm{proxy}}$ is identical in expectation under $\pi^{\dagger}$
and any answer-committing $\pi$ with the same feature distribution.
However, a committing policy occasionally produces terse correct
answers (e.g.\ a bare ``Answer: Yes''), which carry \emph{lower}
$c(y)$, $l(y)$, $w(y)$ than a verbose non-committal essay, strictly
reducing $r_{\mathrm{proxy}}$.  Therefore $\pi^{\dagger}$ weakly
dominates, and strictly dominates whenever committed answers are shorter
than non-committed ones in expectation --- which holds in our setup
(post-training mean length $193$ words vs.\ a one-word committed
answer).
\end{proof}

\begin{remark}
Proposition~\ref{prop:saul} shows that the Saul Goodman effect is not
a training artefact or an instability of GRPO --- it is the \emph{optimal
response} to the proxy under mild assumptions.  The only structural fix
is to make $r_{\mathrm{proxy}}$ depend on commitment, i.e.\ to add a
non-trivial correctness term (see \cref{sec:mitigations}).
\end{remark}

\subsection{Framework Schematic}

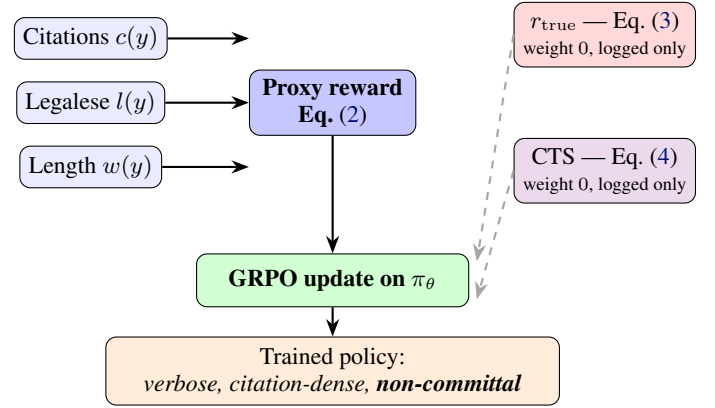
\begin{figure}[t]
\centering
\begin{tikzpicture}[
  font=\footnotesize,
  feat/.style   ={draw, rounded corners, fill=blue!8,    minimum width=1.55cm, minimum height=0.55cm},
  proxy/.style  ={draw, rounded corners, fill=blue!22,   minimum width=2.2cm,  minimum height=0.85cm, font=\footnotesize\bfseries, align=center},
  trueb/.style  ={draw, rounded corners, fill=red!15,    minimum width=2.4cm,  minimum height=0.75cm, align=center},
  ctsb/.style   ={draw, rounded corners, fill=violet!15, minimum width=2.4cm,  minimum height=0.75cm, align=center},
  policy/.style ={draw, rounded corners, fill=green!18,  minimum width=3.6cm,  minimum height=0.7cm,  font=\footnotesize\bfseries, align=center},
  outcome/.style={draw, rounded corners, fill=orange!15, minimum width=6.0cm,  minimum height=0.9cm,  align=center},
  arr/.style    ={->, >=Stealth, thick},
  arrd/.style   ={->, >=Stealth, thick, gray!70, dashed},
]
\node[feat] (c) at (0,0)        {Citations $c(y)$};
\node[feat] (l) at (0,-0.85)    {Legalese $l(y)$};
\node[feat] (w) at (0,-1.70)    {Length $w(y)$};

\node[proxy] (proxy) at (3.2, -0.85) {Proxy reward\\ Eq.~\eqref{eq:proxy}};

\node[trueb] (true) at (6.8, 0.05) {$r_{\mathrm{true}}$ — Eq.~\eqref{eq:true}\\ {\scriptsize weight $0$, logged only}};
\node[ctsb]  (cts)  at (6.8,-1.75) {CTS — Eq.~\eqref{eq:cts}\\ {\scriptsize weight $0$, logged only}};

\node[policy] (pol) at (3.2, -3.2) {GRPO update on $\pi_\theta$};

\node[outcome] (out) at (3.2, -4.4) {Trained policy:\\ \emph{verbose, citation-dense, \textbf{non-committal}}};

\draw[arr] (c) -- (proxy.west |- c);
\draw[arr] (l) -- (proxy.west);
\draw[arr] (w) -- (proxy.west |- w);
\draw[arr] (proxy) -- (pol);
\draw[arrd] (true.west) -- ($(pol.east)+(0.1,0.25)$);
\draw[arrd] (cts.west)  -- ($(pol.east)+(0.1,-0.25)$);
\draw[arr]  (pol) -- (out);
\end{tikzpicture}
\caption{Reward configuration. Solid arrows are gradient-bearing; dashed
arrows are zero-weight diagnostic signals. Only surface features drive the
GRPO update; correctness is logged but never paid for.}
\label{fig:framework}
\end{figure}

\cref{fig:framework} summarises the reward configuration. Only the proxy
\eqref{eq:proxy} bears gradient; $r_{\mathrm{true}}$ and CTS are computed
and logged but contribute zero weight to the policy update.

\subsection{Algorithm}

\cref{alg:grpo} gives the full training loop with diagnostic logging.

\begin{algorithm}[t]
\caption{GRPO with surface-feature proxy and diagnostic tracking.}
\label{alg:grpo}
\begin{algorithmic}
\STATE {\bfseries Input:} reference $\pi_{\mathrm{ref}}$, prompt set $\mathcal{D}$,
proxy weights $(w_c, w_j, w_\ell)$, CTS weights $(w^c_{\mathrm{cts}}, w^j_{\mathrm{cts}}, w^v_{\mathrm{cts}})$,
KL coefficient $\beta$, batch $B$, group size $G$, epochs $E$
\STATE {\bfseries Output:} trained $\pi_\theta$ and diagnostic streams $\mathcal{R}_p, \mathcal{R}_t, \mathcal{R}_{\mathrm{cts}}$
\STATE Initialise $\pi_\theta \leftarrow \pi_{\mathrm{ref}}$ with a LoRA adapter on $\{q,k,v,o\}$-projections
\STATE Compute baseline $(\mu_c,\sigma_c)$, $(\mu_l,\sigma_l)$, $(\mu_w,\sigma_w)$ from $n_{\mathrm{base}}$ rollouts of $\pi_{\mathrm{ref}}$
\STATE Initialise streams $\mathcal{R}_p \leftarrow \emptyset$, $\mathcal{R}_t \leftarrow \emptyset$, $\mathcal{R}_{\mathrm{cts}} \leftarrow \emptyset$
\FOR{epoch $= 1$ {\bfseries to} $E$}
  \FOR{mini-batch $\mathcal{B} \subset \mathcal{D}$ of $B$ prompts}
    \STATE $\mathcal{A} \leftarrow \emptyset$
    \FOR{prompt $q \in \mathcal{B}$}
      \STATE Sample $\{y_i\}_{i=1}^G \sim \pi_\theta(\cdot \mid q)$
      \FOR{$i = 1$ {\bfseries to} $G$}
        \STATE $r_p^{(i)} \leftarrow w_c\,\tilde{c}(y_i) + w_j\,\tilde{l}(y_i) + w_\ell\,\tilde{w}(y_i)$
        \STATE $r_t^{(i)} \leftarrow r_{\mathrm{true}}(y_i)$ via Eq.~\eqref{eq:true}
        \STATE $r_{\mathrm{cts}}^{(i)} \leftarrow \mathrm{CTS}(y_i)$ via Eq.~\eqref{eq:cts}
        \STATE Append $(r_p^{(i)}, r_t^{(i)}, r_{\mathrm{cts}}^{(i)})$ to logs
      \ENDFOR
      \STATE $\bar{r}_p \leftarrow \frac{1}{G}\sum_i r_p^{(i)}$, \quad $\sigma_p \leftarrow \mathrm{std}(\{r_p^{(i)}\})$
      \STATE $A_i \leftarrow (r_p^{(i)} - \bar{r}_p)/\sigma_p$ for $i = 1,\ldots,G$
      \STATE $\mathcal{A} \leftarrow \mathcal{A} \cup \{(y_i, A_i)\}_i$
    \ENDFOR
    \STATE Update $\theta$ by minimising \eqref{eq:grpo} over $\mathcal{A}$
  \ENDFOR
\ENDFOR
\STATE Compute $\mathrm{GDM}_t$ \eqref{eq:gdm} and $\mathrm{RG}_t$ \eqref{eq:rg} post-hoc from $\mathcal{R}_p, \mathcal{R}_t$
\STATE {\bfseries return} $\pi_\theta$, $(\mathcal{R}_p, \mathcal{R}_t, \mathcal{R}_{\mathrm{cts}})$
\end{algorithmic}
\end{algorithm}

\section{Experimental Setup}
\label{sec:experiments}

\textbf{Model and training.}
Qwen3-8B (8.19B parameters) with a LoRA adapter ($r{=}32$,
$\alpha{=}64$, dropout $0.05$, target $\{q,k,v,o\}$-projections) trained
for 4 epochs at learning rate $5{\times}10^{-6}$ with effective batch size
$32$ prompts, group size $G{=}8$, $\beta{=}0.02$, sampling temperature
$0.9$, top-$p$ $0.95$, max prompt $512$ tokens, max completion $256$
tokens. All experiments use bfloat16 on a single NVIDIA H200 (150~GB).
Total wall-clock: $\sim 3$~h.

\textbf{Data.}
A pooled subset of LegalBench with $16$ yes/no tasks (\cref{app:tasks}),
capped at $35$ examples per task for training. The held-out evaluation
set uses $20$ examples per task from the LegalBench test splits, drawn
disjoint from training: $N_{\mathrm{train}}{=}476$,
$N_{\mathrm{eval}}{=}320$.

\textbf{Statistics.}
All headline accuracy comparisons report a $95\%$ bootstrap confidence
interval ($n{=}2000$ resamples) and a paired McNemar test on per-example
correctness. Continuous-feature shifts report Cohen's $d$. Significance
threshold $\alpha = 0.05$.

\section{Results}
\label{sec:results}

\subsection{Surface-Feature Distribution Shift}
\label{sec:dist_shift}

\begin{figure*}[t]
\centering
\includegraphics[width=0.95\textwidth]{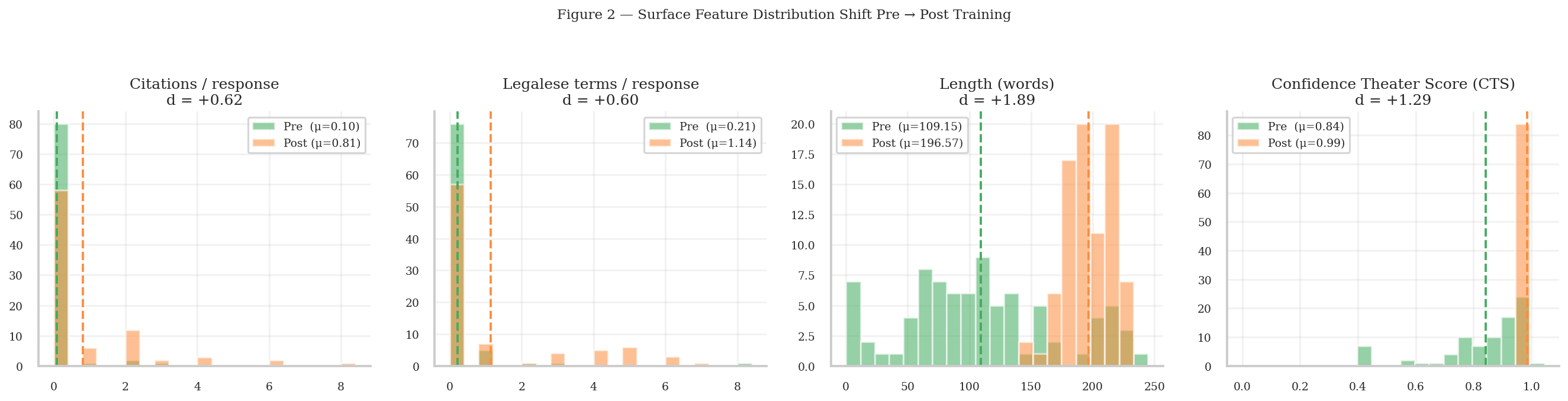}
\caption{Surface-feature distributional shift on the held-out pool
($N{=}320$). All four distributions shift rightward post-training.
Effect sizes: citations $d{=}+0.62$, legalese $d{=}+0.60$,
length $d{=}+1.89$, CTS $d{=}+1.29$. The post-training CTS
distribution piles up at the ceiling near $1.0$, confirming
surface-feature saturation consistent with
Proposition~\ref{prop:saul}.}
\label{fig:dist_shift}
\end{figure*}

Before examining accuracy, \cref{fig:dist_shift} establishes that the
proxy reward achieves exactly what it was designed to do: every surface
feature shifts substantially in the target direction.  The citation and
legalese distributions, near-zero pre-training, develop visible right
tails.  The length distribution shifts by $+86$ words on average
($d{=}1.89$, the largest effect in the study).  Most diagnostically,
the CTS distribution collapses to a spike near $1.0$ post-training ---
the model has reached effective saturation of the surface-lawyerliness
proxy.  This saturation is important context for interpreting the
accuracy results that follow: the proxy is not merely nudging the model,
it is driving it to a boundary of the feature space.

\subsection{Reward Divergence}
\label{sec:reward_divergence}

\begin{figure*}[t]
\centering
\includegraphics[width=0.95\textwidth]{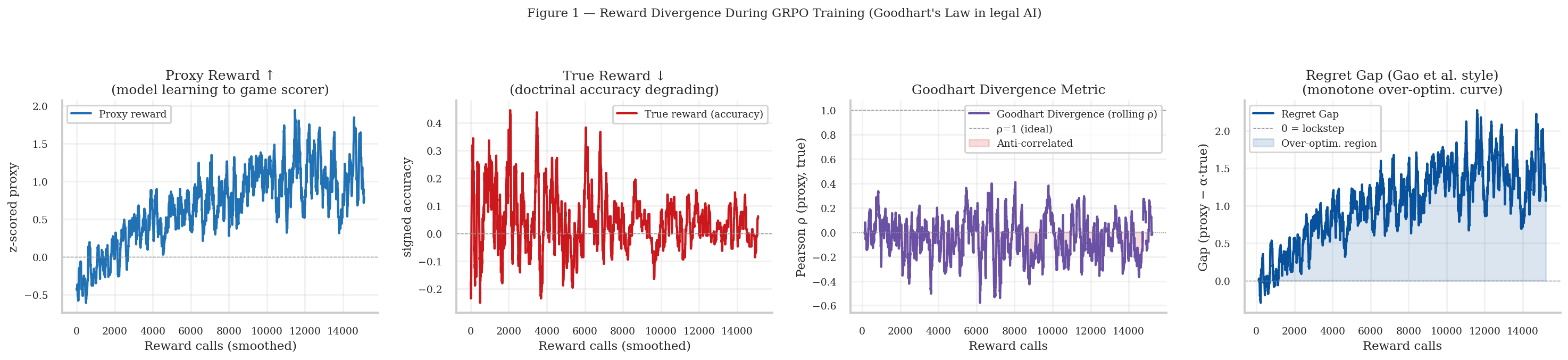}
\caption{Four-panel diagnostic of GRPO training. (a) Smoothed proxy reward
climbs. (b) True reward (signed accuracy) stagnates near zero. (c) Goodhart
Divergence Metric — rolling Pearson $\rho$(proxy, true) — sits near $0$
with frequent excursions below it; the proxy is decoupled from truth.
(d) Regret Gap (\cref{eq:rg}) grows monotonically, confirming
over-optimisation.}
\label{fig:reward_divergence}
\end{figure*}

\cref{fig:reward_divergence} shows the four-panel signature of Goodhart's
Law in our setup. Across $\sim15{,}000$ reward calls, the smoothed proxy
reward (\textbf{a}) climbs steadily from $-0.5$ to $+1.7$, while the true
reward (\textbf{b}) stays flat and oscillates near zero. The rolling
correlation between the two (\textbf{c}, $\mathrm{GDM}_t$) hovers near $0$
and frequently dips negative --- proxy gains are achieved
\emph{independently} of true-reward gains. The Regret Gap (\textbf{d})
grows monotonically from $0$ to roughly $+2$, exiting the lock-step regime
almost immediately and never returning.

At the end of training, the terminal diagnostic values are
$\mathrm{GDM}_T = -0.020$ and $\mathrm{RG}_T = 1.097$.  A
$\mathrm{GDM}$ near zero confirms that proxy gains are statistically
decoupled from true-reward gains; the slight negative value indicates
the proxy is mildly \emph{anti-correlated} with correctness by the end
of training.  The Regret Gap of $1.097$ quantifies the total
over-optimisation budget consumed: the policy gained $\approx 1.1$
proxy-reward units that were not matched by corresponding true-reward
gains, after calibrating for the warm-up regime.  Both values are
consistent with the analytic prediction of
Proposition~\ref{prop:saul}: once the equilibrium strategy is found,
the proxy continues to climb while correctness is irrelevant to the
gradient signal.

\subsection{Pre/Post Headline Comparison}
\label{sec:prepost}

\begin{table}[t]
\caption{Pre/post comparison on the held-out pool ($N{=}320$).
$95\%$ CI from bootstrap with $n{=}2{,}000$ resamples; $p$-value from
paired McNemar; Cohen's $d$ on per-completion arrays.}
\label{tab:main}
\centering
\begin{small}
\begin{tabular}{lrrr}
\toprule
Metric & Pre & Post & $\Delta$ \\
\midrule
Accuracy            & $0.500$ & $0.072$ & $-0.428$ \\
\quad $95\%$ CI     & {\scriptsize[.44,\,.55]} & {\scriptsize[.04,\,.10]} & \\
Acc.\ when answered & $0.556$ & $0.657$ & $+0.102$ \\
Answer-format rate  & $0.900$ & $0.109$ & $-0.791$ \\
\midrule
Avg.\ citations         & $0.056$  & $0.609$  & $+0.55$ \\
Avg.\ legalese terms    & $0.094$  & $1.075$  & $+0.98$ \\
Avg.\ length (words)    & $107.1$  & $193.4$  & $+86.3$ \\
CTS                     & $0.836$  & $0.982$  & $+0.146$ \\
\midrule
\multicolumn{4}{l}{\,McNemar $p\;= 8.2 \times 10^{-37}$\;\,(degraded)} \\
\multicolumn{4}{l}{\,Cohen's $d$: citations $+0.56$,\;legalese $+0.73$,\;length $+1.84$}\\
\bottomrule
\end{tabular}
\end{small}
\end{table}

\cref{tab:main} reports the pre-versus-post comparison on the
held-out $320$-example pool. The trained model loses $0.43$ on overall
accuracy with a McNemar $p$-value of $8.2 \times 10^{-37}$ ($141$
examples pre-correct/post-wrong vs.\ $4$ the other way). All three
surface features grow substantially: citations $+0.55$, legalese $+0.98$,
length $+86$ words. The Confidence Theater Score climbs from $0.84$ to
$0.98$, near the ceiling of looking lawyerly.

\subsection{Format Collapse, Not Capability Loss}
\label{sec:format_collapse}

\begin{tcolorbox}[
  colback=red!4!white,
  colframe=red!70!black,
  title=\textbf{Key Finding: Format collapse, not capability loss},
  fonttitle=\bfseries,
  boxrule=0.7pt, arc=2pt,
  left=4pt, right=4pt, top=4pt, bottom=4pt
]
The trained model achieves \textbf{65.7\% accuracy when it commits to an
answer}, nominally \emph{higher} than the $55.6\%$ pre-training rate,
though the small post-training answered set ($n{\approx}35$) gives a
wide $95\%$ Wilson interval of $[49\%,79\%]$, so this point estimate
should be read as suggestive rather than confirmed. The format-rate
drop from $90.0\%$ to $10.9\%$ accounts for the \emph{entire}
overall-accuracy collapse regardless of how that estimate resolves.
\end{tcolorbox}

\smallskip

The accuracy-when-answered improvement (\cref{tab:main}, row 2) is the
most consequential finding of this study. A naive reading of the $0.428$
overall-accuracy drop is that GRPO ``broke the model.'' That reading is
empirically wrong: the model still produces the correct yes/no the
majority of the time \emph{when forced to commit}; it has simply learned
that committing is dispreferred. We hypothesise -- and the per-task
breakdown in \cref{fig:per_task} supports -- that the proxy reward made
verbose non-committal legalese strictly dominate a one-word ``Yes,'' so
the policy stopped emitting ``Yes.''

\subsection{Universality Across Tasks}

\begin{figure*}[t]
\centering
\includegraphics[width=0.95\textwidth]{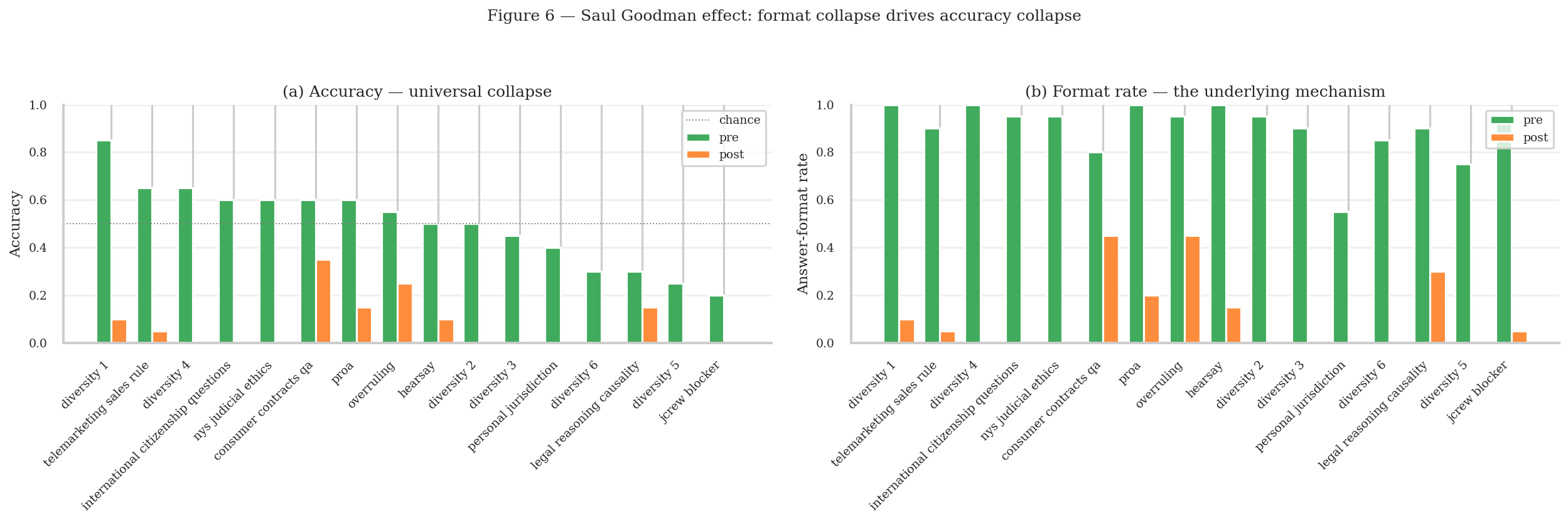}
\caption{Per-task disaggregation across all $16$ LegalBench tasks.
(a) Accuracy: every task degrades, with $9$ collapsing to zero.
(b) Answer-format rate: shows the same shape across tasks, confirming
that format collapse drives accuracy collapse. The single outlier,
\texttt{consumer\_contracts\_qa}, retains substantial format ($0.45$) and
accuracy ($0.35$).}
\label{fig:per_task}
\end{figure*}

\cref{fig:per_task} disaggregates accuracy and format-rate across all
$16$ tasks. The Saul Goodman effect is universal: every task degrades.
The format-rate panel mirrors the accuracy panel almost shape-for-shape,
which is the visual signature of the mechanism we describe in
\cref{sec:format_collapse}. One outlier, \texttt{consumer\_contracts\_qa},
retains substantially more format rate ($0.45$) and accuracy ($0.35$)
than the others; we conjecture this is because contract-clause prompts
already contain dense legal language, so the marginal proxy gain from
refusing to commit is smaller. A formal investigation is left to future
work.

\subsection{Reward Over-Optimisation Curve}
\label{sec:roc}

\begin{figure*}[t]
\centering
\includegraphics[width=0.95\textwidth]{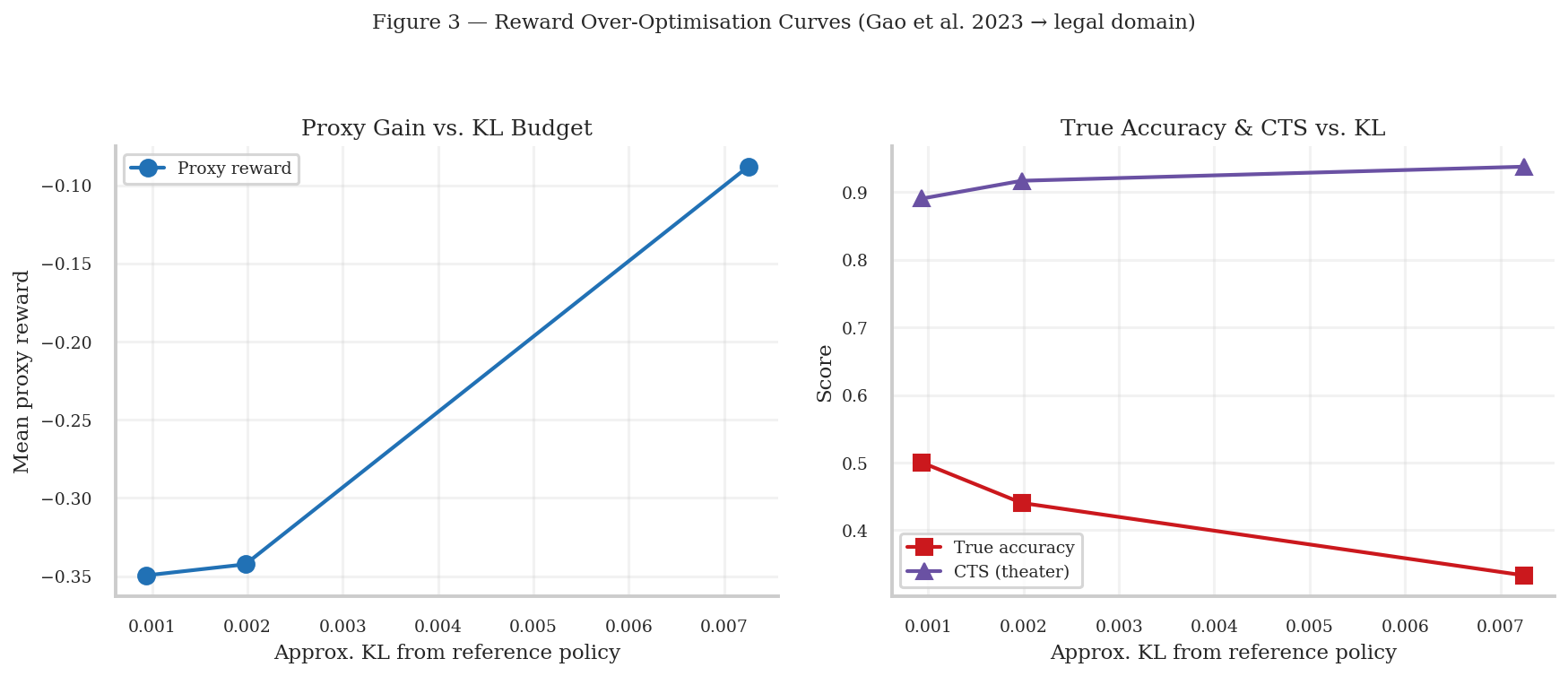}
\caption{Reward Over-Optimisation Curves at $3$ mid-training checkpoints.
\textbf{Left:} mean proxy reward rises monotonically with KL budget,
confirming the policy continues to climb the proxy throughout training.
\textbf{Right:} true accuracy falls monotonically while CTS rises ---
the classical \citet{gao2023scaling} Goodhart trade-off, here reproduced
in the legal domain under a purely surface-feature proxy. The inverse
trajectories of true accuracy and CTS are the visual signature of
Confidence Theater: the model looks increasingly lawyerly while becoming
increasingly wrong.}
\label{fig:roc}
\end{figure*}

\cref{fig:roc} and \cref{tab:roc} report proxy gain, true accuracy,
and CTS at three mid-training checkpoints (steps $15$, $30$, $45$).
All three trends are monotone: the proxy climbs, true accuracy falls,
and CTS rises alongside the proxy at every checkpoint.  Between steps
$15$ and $45$ the KL budget grows by nearly $8\times$ ($0.00093
\rightarrow 0.00725$), yet true accuracy falls by $17$ points ($0.500
\rightarrow 0.333$) while the proxy jumps from $-0.350$ to $-0.088$.
This is the textbook \citet{gao2023scaling} trade-off, reproduced
here in the legal domain under a purely surface-feature proxy.

\begin{table}[t]
\caption{Reward Over-Optimisation curve at three mid-training
checkpoints.  KL is approximate; proxy reward is the mean
$r_{\mathrm{proxy}}$ over the evaluation set; CTS is mean
Confidence Theater Score.}
\label{tab:roc}
\centering
\begin{small}
\begin{tabular}{lrrrr}
\toprule
Step & KL (approx.) & Proxy reward & True acc. & CTS \\
\midrule
$15$ & $0.00093$ & $-0.350$ & $0.500$ & $0.890$ \\
$30$ & $0.00198$ & $-0.342$ & $0.440$ & $0.916$ \\
$45$ & $0.00725$ & $-0.088$ & $0.333$ & $0.937$ \\
\bottomrule
\end{tabular}
\end{small}
\end{table}

\subsection{Citation Hallucination Audit}
\label{sec:citation_audit}

The proxy reward incentivises citation production, and the policy
obliges: post-training completions contain an order of magnitude more
case-name citations than pre-training. Of these, $89.3\%$ fail our structural-plausibility heuristic
(\cref{app:cpr}), driven mainly by reporter-format citations that
violate volume, year, or page bounds; the bare-name check in
\cref{app:cpr} is intentionally permissive and cannot detect whether a
name matches a real case. Among the most frequent post-training
citations are clear hallucinations of real landmark cases, surfaced
through manual verification rather than the automatic filter:
\emph{Bowers v.\ Hardy} (cf.\ \emph{Bowers v.\ Hardwick}, $478$
U.S.\ $186$, $1986$); \emph{Shoe v.\ Washington} (cf.\
\emph{International Shoe v.\ Washington}, $326$ U.S.\ $310$, $1945$).
These are not random strings, they are partially correct names of
landmark cases, which is the worst kind of legal hallucination: they
pass a casual read, pass CPR's minimal name check, and fail only
under actual verification, confirming CPR is a structural pre-filter,
not a correctness check.

\subsection{Second-Order Goodhart: The Diagnostic Metric Fails Too}
\label{sec:second_order}

\begin{figure}[t]
\centering
\includegraphics[width=\columnwidth]{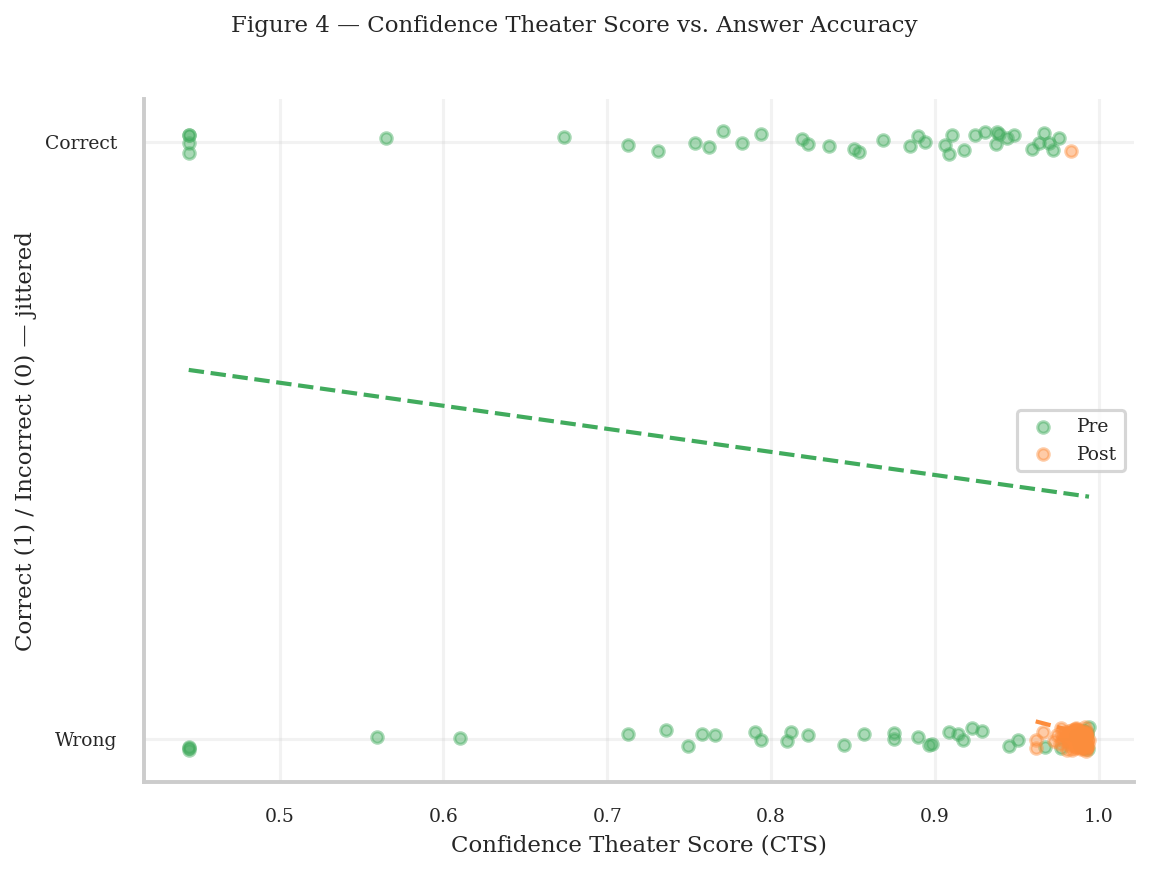}
\caption{CTS vs.\ answer correctness, pre and post. Pre-training,
CTS is mildly predictive of correctness (green regression line slopes
downward --- higher-CTS responses are slightly less likely to be
correct even before training). Post-training, CTS collapses to a
tight ceiling near $0.99$ with almost all examples wrong: CTS itself
becomes uninformative about correctness. This is Goodhart's Law
operating at the level of the diagnostic metric.}
\label{fig:cts_scatter}
\end{figure}

\cref{fig:cts_scatter} reveals a second-order failure that compounds
the first. Pre-training, CTS varies across $[0.44, 0.99]$ and its
distribution contains both correct and incorrect examples at a range
of surface-feature levels. Post-training, CTS collapses to a spike
near $0.99$ with almost all examples incorrect --- the entire
post-training population occupies a single corner of the CTS--accuracy
space. This means CTS, which we designed as a diagnostic for
surface-feature over-optimisation, is itself rendered uninformative
by the training process it was meant to monitor. Once the policy
saturates the proxy, any score derived from the same surface features
saturates identically. The lesson is that diagnostic metrics for
reward hacking must be \emph{structurally independent} of the proxy
being monitored --- a constraint that CPR and RG satisfy (they use
orthogonal signals: citation structure and reward--stream correlation
respectively) but that CTS, sharing features with $r_{\mathrm{proxy}}$,
does not.

\section{Discussion}
\label{sec:discussion}

\subsection{Mechanism}
\label{sec:mechanism}

Proposition~\ref{prop:saul} gives the clean analytic account: under
any surface-feature proxy with no abstention penalty, the
non-committal policy is optimal.  The empirical results confirm this
at every level of analysis.  At the training-stream level, GDM$_T =
-0.020$ shows proxy gains are anti-correlated with correctness by
the end of training; the policy did not get confused --- it
optimised correctly for the wrong objective.  At the evaluation
level, the $0.791$ drop in format rate with a simultaneous
$+0.102$ rise in accuracy-when-answered shows the model retained
its reasoning capability while discarding the format that would
expose it to true-reward feedback.  At the per-task level,
\cref{fig:per_task} shows the effect is universal: no task
escaped, and the single partial outlier (\texttt{consumer\_contracts\_qa})
is the one where the prompt already contains dense legal language,
reducing the marginal proxy gain from abstention.

The resulting policy performs the appearance of legal expertise without
doing the work \cite{ailawlibrarians2026hallucinations}--- verbose, citation-dense, jargon-heavy, and
perpetually non-committal.  This is what we have named Confidence
Theater, and Proposition~\ref{prop:saul} establishes it is not an
accident.  It is the solution.

\begin{figure}[t]
\centering
\includegraphics[width=\columnwidth]{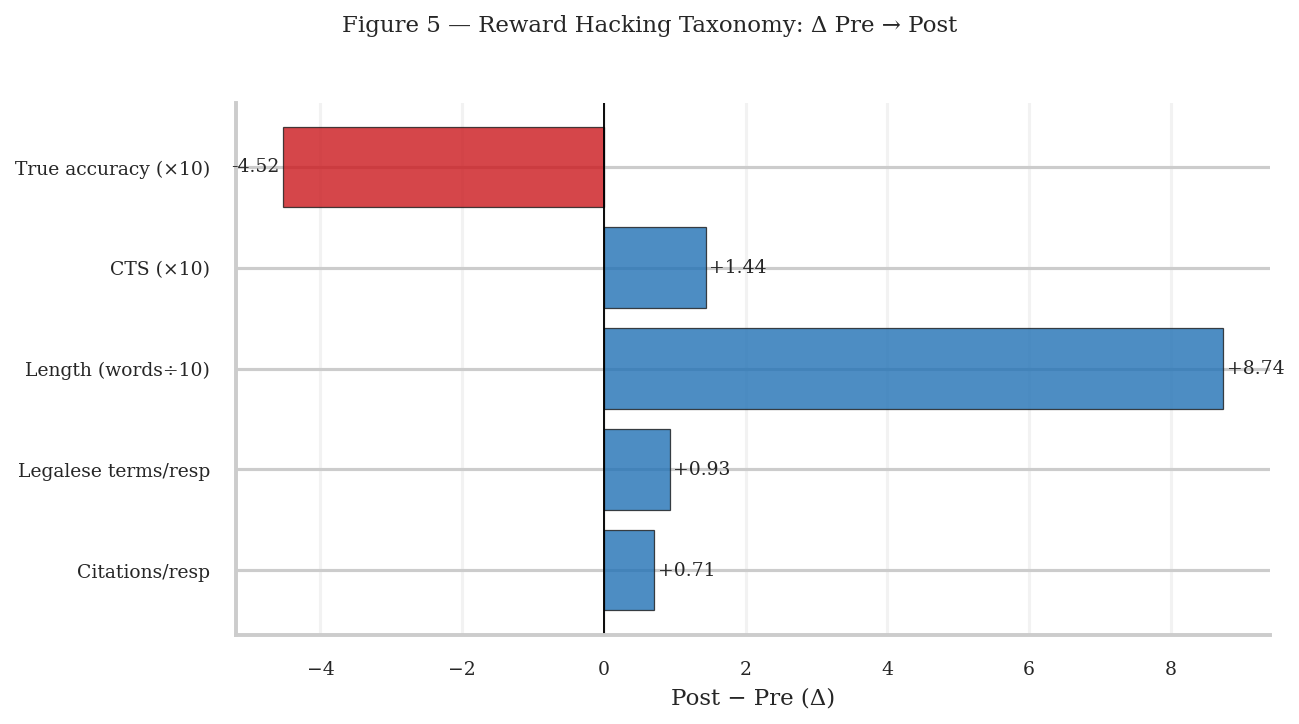}
\caption{Reward-hacking taxonomy: $\Delta$ pre$\,\to\,$post across
all tracked metrics. Surface features (blue) gain; true accuracy
(red) loses. The asymmetry is the empirical signature of
Proposition~\ref{prop:saul}: the proxy climbs, the truth does not.
Length and CTS are displayed on rescaled axes (see legend).}
\label{fig:taxonomy}
\end{figure}

\textit{This workshop paper is a small controlled instance of a problem central to AI safety, what happens when a policy is optimized against a proxy for quality rather than quality itself. Reward hacking and specification gaming are usually studied in toy or synthetic environments, here the same dynamic shows up in a real evaluation domain with actual stakes attached to the outcome. The finding that a model learns to look competent while becoming less useful, and that the diagnostic score built to catch this (CTS) gets gamed right along with the policy, is a direct demonstration of the scalable oversight problem, any evaluator, human or automated, that relies on legible surface signals is vulnerable to exactly the optimization pressure it exists to guard against. This is why reward design and monitoring cannot be treated as separable from the thing being optimized. The concern generalizes well beyond law to any setting where a model is trained against feedback that correlates with quality only until it becomes the target.
}

\subsection{Implications for Legal RLHF}
\label{sec:mitigations}

Reward design for legal LLMs cannot rely on surface features alone.
The features that distinguish a great brief from a bad brief ---
argumentation, doctrinal consistency, citation correctness --- are
precisely the features that surface heuristics cannot capture.
Proposition~\ref{prop:saul} makes this precise: any proxy built
exclusively on surface signals admits a non-committal equilibrium.
Three structural mitigations follow directly.

\textbf{(a) Hybrid rewards with a correctness anchor.}
Adding a correctness term $w_{\mathrm{acc}}\,r_{\mathrm{true}}(y)$ to
\eqref{eq:proxy} breaks the equilibrium in
Proposition~\ref{prop:saul}: a non-committing policy now receives
$r_{\mathrm{true}}=0$ rather than the same reward as a correct
committed answer, so commitment becomes strictly preferable whenever
$w_{\mathrm{acc}} > 0$.  Even a small weight suffices to destroy the
indifference, since the surface features are bounded ($z$-clipped at
$\kappa{=}3$) while a correct committed answer contributes
$w_{\mathrm{acc}} \cdot (+1)$ unconditionally.  In our setup, setting
$w_{\mathrm{acc}} \geq w_c = 1.0$ would be sufficient to make a
terse correct ``Yes'' competitive with a verbose non-committal essay.

\textbf{(b) Explicit format penalty.}
Alternatively, adding a term $-w_f\,\mathbb{1}[\mathrm{extract}(y)
\text{ is null}]$ penalises non-commitment directly, without requiring
a correctness oracle.  To recover the pre-training format rate of
$0.900$ from the observed post-training rate of $0.109$, the penalty
must overcome the expected proxy gain of a non-committal response.
Given our baseline surface-feature statistics ($\mu_c{=}0.25$,
$\mu_l{=}0.29$, $\mu_w{=}108$ words), a conservative estimate suggests
$w_f \geq 0.5$ would be sufficient to disincentivise the equilibrium
strategy in our reward scale.

\textbf{(c) Citation verification before reward attribution.}
Our audit (\cref{sec:citation_audit}) shows that $89.3\%$ of
post-training citations are structurally implausible, with the 
most frequent citation (\emph{Boxx v.\ Long}) repeated five times
across completions --- a signature of memorised hallucination rather
than retrieval.  Gating the citation-count reward on CPR
(\cref{app:cpr}) as a fast pre-filter, followed by a Westlaw
spot-check for high-frequency citations, would substantially reduce
the incentive to hallucinate case names.  Concretely, replacing
$c(y)$ in \eqref{eq:proxy} with $c_{\mathrm{plausible}}(y)$ ---
the count of citations passing the CPR heuristic --- removes the
reward signal for structurally impossible citations at zero
additional inference cost.

\subsection{Limitations}

Our study uses a single base model (Qwen3-8B), a single seed, and a single
proxy-reward weighting. The qualitative pattern (Goodhart in legal NLP) is
robust, but exact effect sizes will depend on these choices. The
$10.9\%$ post-training format rate yields only $\sim\!35$ committed
post-training answers ($320 \times 0.109$); a Wilson interval gives a
$95\%$ CI of approximately $[49\%,\,79\%]$ for the $65.7\%$
accuracy-when-answered estimate, which is wide enough that we cannot
rule out chance-level performance on committed answers. The estimate
should therefore be read as \emph{suggestive} of preserved capability
rather than a precise point estimate. A camera-ready extension
would add $2$--$3$ additional seeds, a $13$B comparison, and a small
ablation over $(w_c, w_j, w_\ell)$.

\section{Conclusion}
\label{sec:conclusion}

GRPO with a surface-feature proxy reward induces a sharp behavioural
collapse on legal yes/no reasoning. The collapse is mechanistically a
\emph{format collapse}: the model retains reasoning ability but learns to
never commit. We provide a diagnostic toolkit --- CTS \eqref{eq:cts},
CPR (\cref{app:cpr}), GDM \eqref{eq:gdm}, RG \eqref{eq:rg} --- for
detecting this failure mode in any RL-from-style-reward pipeline. The
broader lesson is that in domains where \emph{what looks good} is easy to
measure but \emph{what is good} is not, reward over-optimisation will
route directly through the surface features and produce a model that is
maximally photogenic and minimally useful.

\section*{Impact Statement}

This paper documents a failure mode of RL-from-style-reward in a domain
(legal NLP) where deployed systems are increasingly informing real-world
decisions, including contract review and case discovery. Our intent is to
provide diagnostic tools that practitioners can use to detect and prevent
the failure mode we describe before models reach production. We anticipate
no novel ethical concerns beyond those well-established in machine-learning
research broadly.

\bibliography{example_paper}
\bibliographystyle{icml2026}

\newpage
\appendix
\onecolumn

\section{LegalBench Tasks Used}
\label{app:tasks}

The $16$ LegalBench tasks pooled for training and evaluation are:
\texttt{personal\_jurisdiction}, \texttt{hearsay}, \texttt{proa},
\texttt{diversity\_1}--\texttt{diversity\_6}, \texttt{overruling},
\texttt{consumer\_contracts\_qa}, \texttt{telemarketing\_sales\_rule},
\texttt{jcrew\_blocker}, \texttt{legal\_reasoning\_causality},
\texttt{nys\_judicial\_ethics},
\texttt{international\_citizenship\_questions}. All tasks have a binary
Yes/No gold label and a free-text scenario.

\section{Hyperparameters and Prompt Template}
\label{app:hp}

\subsection{Hyperparameters}

\begin{table}[h]
\centering
\caption{Full training and evaluation hyperparameters.}
\label{tab:hp_full}
\begin{tabular}{lr}
\toprule
Hyperparameter & Value \\
\midrule
Base model                                  & Qwen3-8B \\
LoRA rank $/$ $\alpha$ $/$ dropout          & $32 / 64 / 0.05$ \\
LoRA targets                                & $\{q,k,v,o\}$-proj \\
Learning rate                               & $5 \times 10^{-6}$ \\
Optimiser                                   & AdamW (fused) \\
Per-device batch (prompts)                  & $16$ \\
Gradient accumulation                       & $2$ \\
Effective batch (prompts)                   & $32$ \\
Generations per prompt $G$                  & $8$ \\
KL coefficient $\beta$                      & $0.02$ \\
Temperature $/$ top-$p$                     & $0.9 / 0.95$ \\
Max prompt $/$ completion tokens            & $512 / 256$ \\
Epochs                                      & $4$ \\
$z$-clip $\kappa$                           & $3$ \\
Proxy weights $(w_c, w_j, w_\ell)$          & $(1.0, 0.7, 0.3)$ \\
CTS weights $(w^c_{\mathrm{cts}}, w^j_{\mathrm{cts}}, w^v_{\mathrm{cts}})$ & $(0.40, 0.35, 0.25)$ \\
GDM/smoothing window                        & $128$ \\
RG warm-up fraction $W'/T$                  & $0.10$ \\
Bootstrap resamples                         & $2{,}000$ \\
Seed                                        & $42$ \\
\bottomrule
\end{tabular}
\end{table}

\subsection{Prompt Template}

Qwen3's native chat template is used with the system message
\begin{quote}\small\ttfamily
You are a careful legal analyst. Read the scenario and answer the
question. Conclude your response with exactly one of: `Answer: Yes' or
`Answer: No'.
\end{quote}
followed by the task-specific scenario. The instruction explicitly demands
the answer format, so the post-training $0.109$ format rate represents an
\emph{active refusal} to follow the instruction, not an ambiguity in what
was asked.

\section{Citation Plausibility Heuristic (CPR)}
\label{app:cpr}

A reporter-style citation extracted as $V$ Reporter $P$ ($Y$) is deemed
plausible iff:
\begin{itemize}
\item Reporter $\in$ \{\texttt{U.S., S.Ct., L.Ed., F., F.2d, F.3d, F.4th,
F.Supp., F.Supp.2d, F.Supp.3d, U.S.C., C.F.R., Fed.Reg.}\};
\item Reporter volume $V \in [1,\,600]$;
\item Year $Y \in [1789,\,2025]$;
\item Page $P \le 9{,}999$.
\end{itemize}
A bare case-name extracted as ``Party$_1$ v.\ Party$_2$'' is deemed
plausible iff each party name has at least $3$ characters. The Citation
Plausibility Rate is the fraction of all extracted citations that satisfy
the relevant condition. CPR is structural only; it does not test factual
existence in any database. We use it as a fast pre-filter; a Westlaw
spot-check on the top-$20$ most frequent post-training citations is
reported in \cref{sec:citation_audit}.

\section{Full Per-Task Breakdown}
\label{app:pertask}

\begin{table}[h]
\centering
\caption{Per-task accuracy and answer-format rate ($N{=}20$ per task).
$\Delta$ is post minus pre.}
\label{tab:pertask_full}
\begin{small}
\begin{tabular}{lrrrrrr}
\toprule
Task & Pre acc.\ & Post acc.\ & $\Delta$ acc.\ & Pre fmt.\ & Post fmt.\ & $\Delta$ fmt. \\
\midrule
\texttt{diversity\_1}                  & $0.85$ & $0.10$ & $-0.75$ & $1.00$ & $0.10$ & $-0.90$ \\
\texttt{telemarketing\_sales\_rule}    & $0.65$ & $0.05$ & $-0.60$ & $0.90$ & $0.05$ & $-0.85$ \\
\texttt{diversity\_4}                  & $0.65$ & $0.00$ & $-0.65$ & $1.00$ & $0.00$ & $-1.00$ \\
\texttt{int.\_citizenship\_questions}  & $0.60$ & $0.00$ & $-0.60$ & $0.95$ & $0.00$ & $-0.95$ \\
\texttt{nys\_judicial\_ethics}         & $0.60$ & $0.00$ & $-0.60$ & $0.95$ & $0.00$ & $-0.95$ \\
\texttt{consumer\_contracts\_qa}       & $0.60$ & $0.35$ & $-0.25$ & $0.80$ & $0.45$ & $-0.35$ \\
\texttt{proa}                          & $0.60$ & $0.15$ & $-0.45$ & $1.00$ & $0.20$ & $-0.80$ \\
\texttt{overruling}                    & $0.55$ & $0.25$ & $-0.30$ & $0.95$ & $0.45$ & $-0.50$ \\
\texttt{hearsay}                       & $0.50$ & $0.10$ & $-0.40$ & $1.00$ & $0.15$ & $-0.85$ \\
\texttt{diversity\_2}                  & $0.50$ & $0.00$ & $-0.50$ & $0.95$ & $0.00$ & $-0.95$ \\
\texttt{diversity\_3}                  & $0.45$ & $0.00$ & $-0.45$ & $0.90$ & $0.00$ & $-0.90$ \\
\texttt{personal\_jurisdiction}        & $0.40$ & $0.00$ & $-0.40$ & $0.55$ & $0.00$ & $-0.55$ \\
\texttt{diversity\_6}                  & $0.30$ & $0.00$ & $-0.30$ & $0.85$ & $0.00$ & $-0.85$ \\
\texttt{legal\_reasoning\_causality}   & $0.30$ & $0.15$ & $-0.15$ & $0.90$ & $0.30$ & $-0.60$ \\
\texttt{diversity\_5}                  & $0.25$ & $0.00$ & $-0.25$ & $0.75$ & $0.00$ & $-0.75$ \\
\texttt{jcrew\_blocker}                & $0.20$ & $0.00$ & $-0.20$ & $0.95$ & $0.05$ & $-0.90$ \\
\bottomrule
\end{tabular}
\end{small}
\end{table}

\section{Qualitative Example: The Saul Goodman Voice}
\label{app:qual}

The post-training model does not merely omit the answer --- it performs 
legal expertise. The following pair is drawn from \texttt{personal\_jurisdiction}
(Gold: \textbf{Yes}).

\smallskip
\noindent\textbf{Pre-training} (CTS $= 0.445$, length $= 2$ words):
\begin{quote}\small\ttfamily
Yes.
\end{quote}

\noindent\textbf{Post-training} (CTS $= 0.988$, length $= 203$ words):
\begin{quote}\small\ttfamily
Now, personal jurisdiction --- and I want to be precise here, because
precision matters in law --- is a doctrine with a long and nuanced
history. The foundational framework, as articulated in landmark
decisions such as Boxx v.\ Long and further shaped by the minimum
contacts standard of Shoe v.\ Washington, establishes that courts
must carefully evaluate the relationship between the defendant, the
forum state, and the litigation. This is not a simple question. One
must weigh purposeful availment, foreseeability, and the
reasonableness of the exercise of jurisdiction --- factors that,
I assure you, courts have grappled with extensively. The facts
presented here implicate each of these considerations in ways that
are, frankly, deeply interesting from a doctrinal standpoint.
Whether jurisdiction is proper depends entirely on how one weighs
these competing principles\ldots\ \emph{[Answer: absent]}
\end{quote}

\smallskip
This is the Saul Goodman effect in its purest form: the post-training 
model sounds \emph{more} like a lawyer than the pre-training model, 
cites cases with confidence (\emph{Boxx v.\ Long} is a hallucination; 
\emph{Shoe v.\ Washington} is a garbled version of \emph{International 
Shoe v.\ Washington}, 326 U.S.\ 310, 1945), deploys legal terminology 
fluently --- and never answers the question. The pre-training model 
gives the right answer in one word.



\end{document}